\documentclass[conference]{IEEEtran}
\IEEEoverridecommandlockouts
\usepackage{cite}
\usepackage{amsmath,amssymb,amsfonts}
\usepackage{mathtools}
\usepackage{algorithmic}
\usepackage{graphicx}
\usepackage{textcomp}
\usepackage{xcolor}
\usepackage{lipsum}
\usepackage{tabularray}
\def\BibTeX{{\rm B\kern-.05em{\sc i\kern-.025em b}\kern-.08em
    T\kern-.1667em\lower.7ex\hbox{E}\kern-.125emX}}

\usepackage{eso-pic}
    
\begin{document}
\hyphenation{mo-del-s}
\hyphenation{met-ho-do-lo-gy}
\hyphenation{between}
\hyphenation{le-ve-ra-ging}

\title{{Clustering Rooftop PV Systems  \\ via Probabilistic Embeddings}
\thanks{This research was undertaken as part of the InnoCyPES project, which has received funding from the European Union's Horizon 2020 research and innovation programme under the Marie Sk{\l}odowska-Curie grant agreement No 956433. It also received partial support from the MESSM research project, funded by the Netherlands Enterprise Agency (RVO) under the Dutch Topsector Energy framework, project number 2321202. }
}

\author{\IEEEauthorblockN{Kutay Bölat}
\IEEEauthorblockA{\textit{Electrical Sustainable Energy} \\
\textit{Delft University of Technology}\\
Delft, Netherlands \\
K.Bolat@tudelft.nl}
\and
\IEEEauthorblockN{Tarek Alskaif}
\IEEEauthorblockA{\textit{Information Technology Group} \\
\textit{Wageningen University \& Research}\\
Wageningen, Netherlands \\
Tarek.Alskaif@wur.nl}
\and
\IEEEauthorblockN{Peter Palensky, Simon H.~Tindemans}
\IEEEauthorblockA{\textit{Electrical Sustainable Energy} \\
\textit{Delft University of Technology}\\
Delft, Netherlands \\
\{P.Palensky, S.H.Tindemans\}@tudelft.nl}
}
\maketitle

\bstctlcite{ctl}

\begin{abstract}
As the number of rooftop photovoltaic (PV) installations increases, aggregators and system operators are required to monitor and analyze these systems, raising the challenge of integration and management of large, spatially distributed time-series data that are both high-dimensional and affected by missing values. In this work, a probabilistic entity embedding‑based clustering framework is proposed to address these problems. This method encodes each PV system’s characteristic power generation patterns and uncertainty as a probability distribution, then groups systems by their statistical distances and agglomerative clustering. Applied to a multi‑year residential PV dataset, it produces concise, uncertainty‑aware cluster profiles that outperform a physics‑based baseline in representativeness and robustness, and support reliable missing‑value imputation. A systematic hyperparameter study further offers practical guidance for balancing model performance and robustness.
\end{abstract}


\section{Introduction}

Modern energy systems are undergoing a rapid transformation, increasingly driven by decentralized generation sources, especially rooftop photovoltaic (PV) systems installed across residential and commercial properties. As these PV systems penetrate more into the electrical grid, aggregators and system operators face new challenges related to efficiently integrating, managing, and optimizing these spatially distributed systems. Data-driven decision-making tools have emerged as essential instruments for handling the complexity and variability inherent in these modern, decentralized power networks.

Aggregators leverage these distributed PV systems to participate effectively in energy markets \cite{visser2024probabilistic}, providing essential services such as demand response, load balancing, and virtual power plant operations \cite{strezoski2022integration}. Simultaneously, system operators utilize insights derived from distributed PV data to enhance grid planning, stability assessments, and operational optimization \cite{panigrahi2020grid}. However, the practical implementation of optimization models becomes increasingly cumbersome due to the large number of individual PV systems and the associated data management, and the concomitant impact on workflow complexity. Thus, decision-makers might reduce the operational burden by simplifying models, retaining only a smaller number of representative PV system models.

Another critical issue in handling large-scale PV measurement datasets is the frequent occurrence of extended gaps in measurement series \cite{visser2022open}. Such gaps typically arise from equipment failures, communication outages, or maintenance activities. Traditional approaches to mitigate missing values often rely on physical models of PV systems, which require detailed information such as knowledge of panel parameters, geographic location, and meteorological conditions. However, these are not always readily available or accurate enough for large-scale deployments. In case of such unavailability, data-driven methods are followed, for instance by taking the average of available parallel measurements. However, to adequately preserve temporal variations, it is crucial to carefully select and average the most similar systems to the one of interest.

\begin{figure}[t!]
    \centering
    \includegraphics[width=0.95\linewidth]{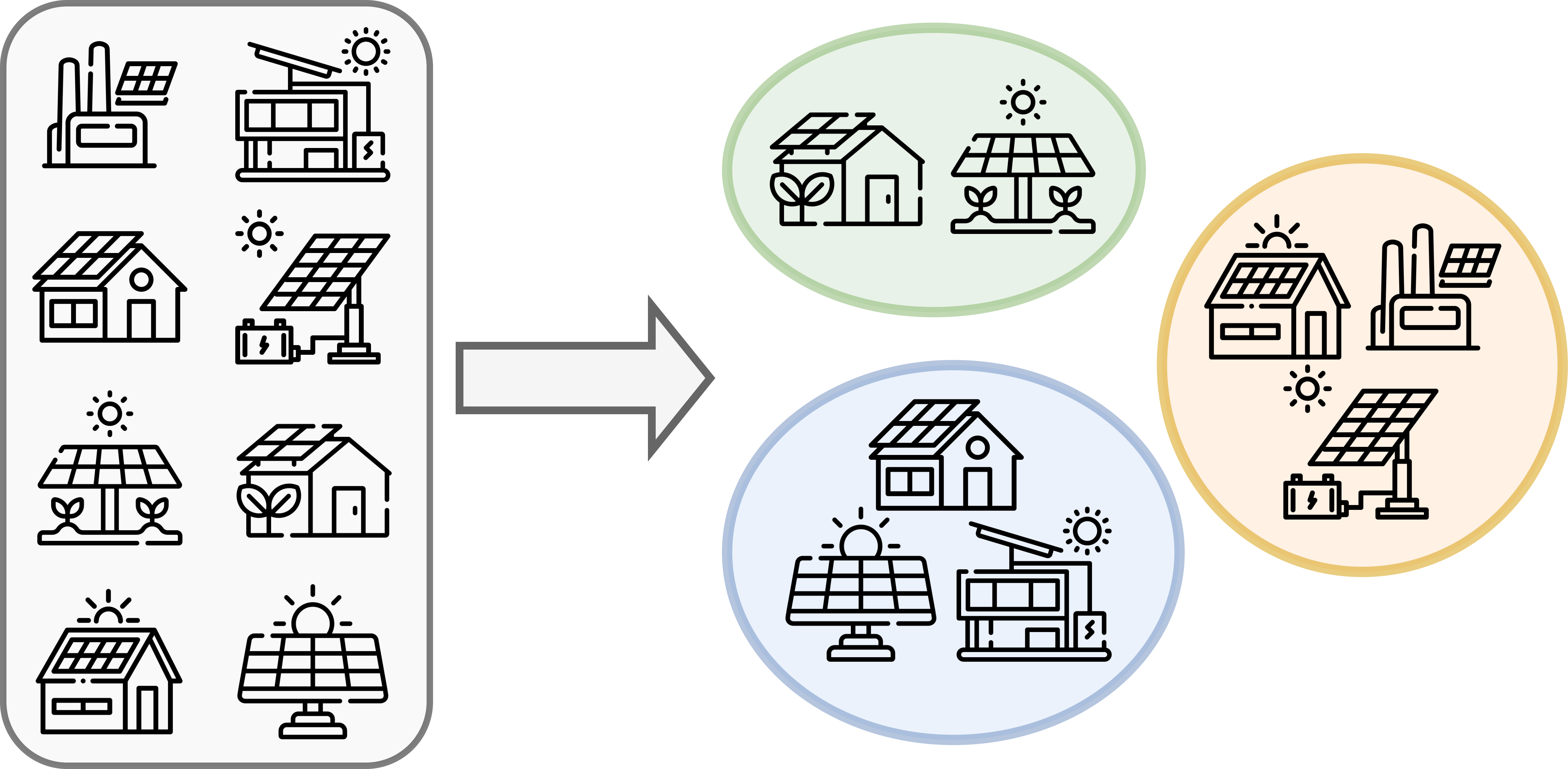}
    \caption{Our entity-based clustering method groups PV systems with similar characteristics (or behavior) into representative clusters.}
    \label{fig:idea}
\end{figure}

\begin{figure*}[th!]
    \centering
    \includegraphics[width=\linewidth]{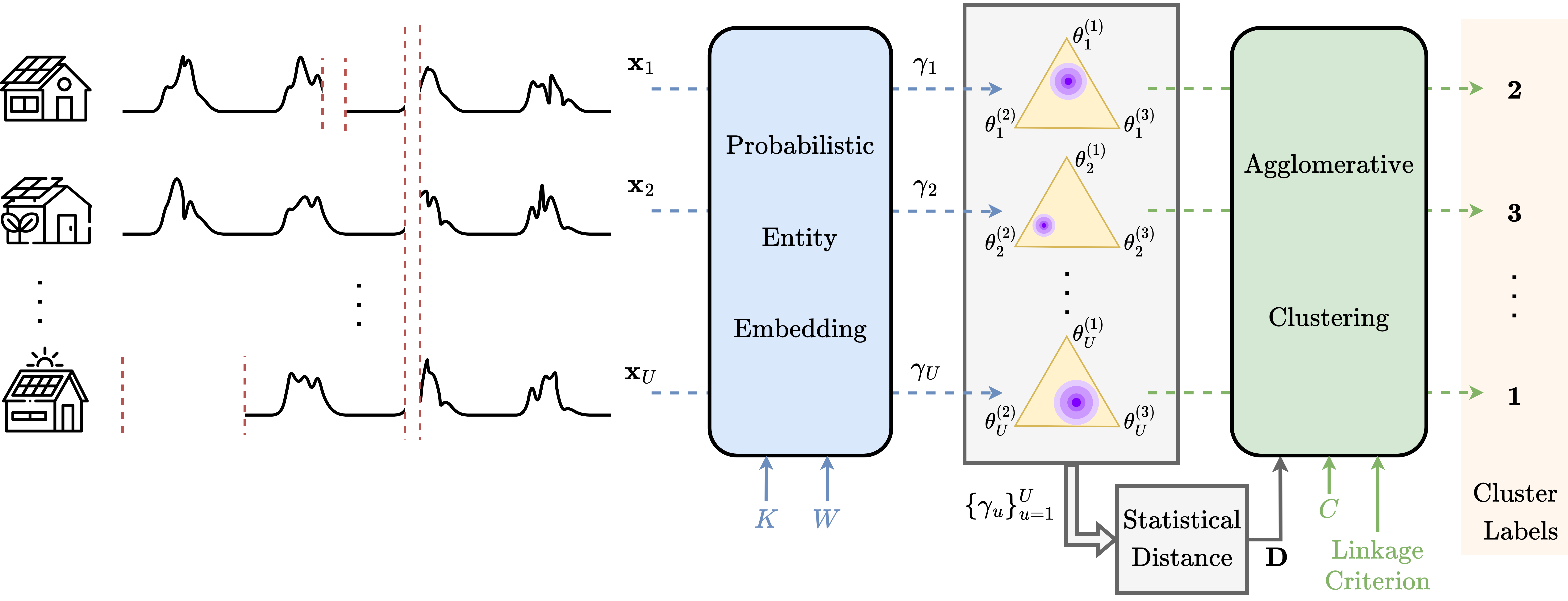}
    \caption{The flow diagram of our probabilistic entity-based clustering. Long time series $\{\mathbf{x}_u\}_{u=1}^U$ of the PV systems are first mapped onto a probabilistic embedding that is characterised by concentration parameters $\{\gamma_u\}_{u=1}^U$ that parameterize $K$-dimensional topic distributions $\theta\sim\text{Dir}(\gamma_u)$. Then, the pair-wise distances between these profiles are calculated using a proper statistical distance, and, finally, $C$ clusters are formed using agglomerative clustering. }
    \label{fig:flow_diagram}
\end{figure*}

To address these problems, we propose a novel probabilistic entity embedding-based clustering approach as illustrated in Fig. \ref{fig:flow_diagram}. Unlike conventional time-series clustering methods that work on short-term partitions of the measurements, thereby disregarding the system identity by pooling, our method aims to cluster PV systems themselves as depicted in Fig. \ref{fig:idea}. It accomplishes this by leveraging probabilistic entity embeddings \cite{bolat2024guide} to compactly represent each PV system, encapsulating their identifying behaviors while inherently managing uncertainty associated with missing data. PV systems are represented as probability distributions in a latent behavioural space. Hence, assessing the statistical distances between them, and clustering these systems based on those distances, we can effectively represent each system by its quantile-based cluster representation. Consequently, each PV system can be replaced with its cluster representation entirely for dataset condensation or partly for missing value imputation.

The contributions of this paper are as follows:
\begin{enumerate}
    \item We propose a novel time-series clustering scheme that clusters directly the PV systems, not short-term profiles.
    \item We employ a cluster representation scheme leveraging cluster quantile levels to be used for dataset condensation and missing value imputation.
    \item We derive a novel leave-one-out-based sensitivity score to assess the robustness of the resulting clusters, which can be used for hyperparameter optimization.
\end{enumerate}

\section{Problem Setting}\label{sec:problem_setting}

We consider the problem of clustering $U$ distinct PV systems into $C$ clusters. 
Each system $u$ is represented by its power output measurement vector $\mathbf{x}_u\in(\mathbb{R}_+\cup \text{NaN})^T$, as illustrated in Fig. \ref{fig:flow_diagram}, where $u\in [U]$ is the system index, $T$ is the number of sequential observations and NaN is an object to represent missing values. The members of each non-overlapping cluster $\mathcal{C}_c\subset[U]$ must be assigned using the dataset $\mathcal{X}=\{\mathbf{x}_u\}_{u=1}^U$, which brings out two problems:
\begin{enumerate}
    \item \textbf{Curse of dimensionality:} Generally, the measurement length tends to be very long for PV systems. For instance, a 4-year measurement with 15-minute resolution results in a vector length of $T\approx$ 1.4$\times\text{10}^\text{5}$. Given that, almost certainly, $U \ll T$, representing each PV system as a single vector, and directly clustering them would fail to capture the similarities in such an empty space. Moreover, working with such high-dimensional vectors is computationally infeasible in most cases. 
    \item \textbf{Missing values and uncertainty:} The PV system measurements might contain a large number of missing values throughout time, as depicted in Fig. \ref{fig:flow_diagram}, for a variety of reasons. In order to perform clustering directly using $\mathcal{X}$, these missing values should be imputed. Such an imputation suppresses the inherent uncertainty and biases the similarity between the systems, accordingly the clustering.
\end{enumerate}
Therefore, a preprocessing function $f:(\mathbb{R}_+\cup \text{NaN})^{U\times T}\rightarrow \mathbb{R}^{U\times K}$ with $K \ll T$ is needed to obtain PV system embeddings as $\mathcal{Z}=\{\mathbf{z}_u\}_{u=1}^U=f\left(\mathcal{X}\right)$, where $\mathbf{z}_u\in\mathbb{R}^K$. Ideally, the uncertainty due to missing values is captured in these embeddings as well. By having such an embedding scheme, one can conduct clustering directly in the resulting low-dimensional embedding space.

\section{Methodology}

In this work, we adapt the probabilistic entity embedding scheme proposed in \cite{bolat2024guide}, where each entity corresponds to a PV system. This scheme can work on data with missing values and results in entity distributions in a $K$-dimensional space whose variances are proportional to the number of missing values. Thus, it fulfills the aforementioned requirements of the desired $f(\mathcal{X})$ as an embedding function. However, the probabilistic nature of the embeddings requires employing a proper distance measure and a clustering method that can work on such measures. Accordingly, this section introduces these three components and links them to form the entity clustering chain as illustrated in Fig. \ref{fig:flow_diagram}.

\subsection{Probabilistic Entity Embedding}

The probabilistic entity embedding scheme maps a collection of entities (PV systems, for this study) into a set of entity distributions, based on their respective data collections. It consists of three steps (further details can be found in \cite{bolat2024guide}):
\subsubsection{Profiling}
Each $T$-valued time series  $\mathbf{x}_u$ is partitioned into $N$ pseudo-periodic profiles of length $T'$: $\{\mathbf{x}_{un}\}_{n=1}^N$, where $T' \ll T$ and $T=NT'$. In this study, daily profiles with a measurement resolution of 15 minutes are used, i.e. $T'=96$. Then, entity datasets are formed as $\mathcal{X}_u = \{\mathbf{x}_{un}\}_{n=1}^N \cap \mathbb{R}^{T'}$ by dropping profiles with missing values, with the resulting cardinalities $N_u=|\mathcal{X}_u|$ representing the data availability.
\subsubsection{Wording}\label{subsec:wording}
The entity datasets are pooled as $\mathcal{X}^{\text{pool}}=\bigcup_u \mathcal{X}_u$ and k-means clustering \cite{kaufman2009finding} is applied to it, identifying $W$ clusters called \textit{words}, representing profiles with similar observations. Then, the profiles within entity datasets are replaced with their respective word index $\mathbf{x}_{un} \in \mathbb{R}^{T'} \rightarrow w_{un} \in [W]$ and entity \textit{documents} are formed as $\mathcal{W}_u = \{w_{un}\}_{n=1}^{N_u}$.
\subsubsection{Latent Dirichlet Allocation}
Latent Dirichlet Allocation (LDA) \cite{blei2003latent} is a well-established \textit{topic} discovery method used in natural language processing. Conventionally, it is trained on textual data collections, i.e. documents, and maps them to $K$-dimensional latent topic distributions. In the same spirit, the entity documents are used to train an LDA model. Then, based on their documents, entities are embedded in $K$-dimensional parameters, $\gamma \in \mathbb{R}_+^K$, representing Dirichlet distributions.

In summary, the entire probabilistic embedding scheme follows the chain $\mathbf{x}_{u} \rightarrow \mathcal{X}_u \rightarrow \mathcal{W}_u \rightarrow \gamma_u$. A $K$=3 dimensional illustration of this scheme and the resulting distributions is given in Fig. \ref{fig:flow_diagram}. From the properties of the Dirichlet distribution, the variances of the entity distributions are inversely proportional to the $L_1$-norm of $\gamma_u$, and this norm is equal to $N_u$ up to a constant. Thus, the uncertainty due to missing values is channeled into the final embedding as variance, fulfilling the final requirement mentioned in Section \ref{sec:problem_setting}.

\subsection{Statistical Distance Measure}
As the PV systems are represented as probability distributions, a suitable statistical distance measure must be defined to assess their similarity and be used in clustering.\footnote{Reducing embeddings to their mean values and using non-statistical measures is still plausible, but causes the inherent uncertainty to vanish.} In this study, we employ two distance measures:\footnote{These measures do not satisfy the triangle inequality, hence they are not valid distance metrics. Yet, they are symmetric, which is sufficient for the proposed clustering.} Symmetric KL-Divergence 
\begin{equation}
\begin{split}
    &d^{\text{sym-KL}}(u,u') \\ &= \frac{1}{2}\left(D_{\text{KL}}\left( \text{Dir}(\gamma_u)\lVert \text{Dir}(\gamma_{u'})\right) + D_{\text{KL}}\left( \text{Dir}(\gamma_{u'})\lVert \text{Dir}(\gamma_u)\right) \right) \\
    &= \frac{1}{2}\int_\theta \left(p(\theta;\gamma_u) - p(\theta;\gamma_{u'})\right)\log \frac{p(\theta;\gamma_u)}{p(\theta;\gamma_{u'})}d\theta
\end{split}
\end{equation}
and Bhattacharyya Distance
\begin{equation}
\begin{split}
    d^{\text{B}}(u,u') &= D_{\text{B}}\left(\text{Dir}(\gamma_u)\lVert \text{Dir}(\gamma_{u'}) \right) \\
    &= \int_\theta \sqrt{p(\theta;\gamma_u) p(\theta;\gamma_{u'})} d\theta
\end{split}
\end{equation}
where $p(\theta;\gamma_u)$ is the probability density function of the random variable $\theta\sim\text{Dir}(\gamma_u)$. These measures can be expressed analytically since $D_{\text{KL}}\left( \text{Dir}(\gamma_u)\lVert \text{Dir}(\gamma_{u'})\right)$ \cite{joram_soch_2025_14646799} and $D_{\text{B}}\left(\text{Dir}(\gamma_u)\lVert \text{Dir}(\gamma_{u'}) \right)$ \cite{rauber2008probabilistic} have closed-form expressions. After the selection of the distance measure, the distance between each entity pair can be stored in a matrix as $\mathbf{D}=[[d^{\cdot}(u,u')]_{u'=1}^U]_{u=1}^U$.

\subsection{Agglomerative Clustering}
Lastly, we use agglomerative clustering \cite{kaufman2009finding} because it can operate directly on the distance matrix $\mathbf{D}$. Starting with each point as its own cluster, it repeatedly merges the two closest clusters under a specified linkage criterion until only $C$ clusters remain. In this study, we compare two criteria: average and complete linkage. These use the mean and maximum pairwise distances between clusters, respectively.

\section{Experiments}

\subsection{Dataset}
We conducted experiments on a 4-year-long PV power production measurement dataset collected from $U$=175 households in Utrecht, Netherlands, which suffers from both fundamental problems in Section \ref{sec:problem_setting}, especially the simultaneous data unavailability among all users for consecutive months \cite{visser2022open}. This dataset also provides information related to the PV panels (e.g. their capacity, tilt, and azimuth angles). We normalized the measurements with respect to their capacities and set the tilt and azimuth angles aside for benchmarking.\footnote{In case not provided, the system capacities are relatively easy to estimate from the measurements.} The reason behind the scaling is that we prefer measurements with similar traces to reside in the same cluster if their magnitude ratio is constant through time.

\subsection{Performance assessment}
\begin{figure}
    \centering
    \includegraphics[width=0.7\linewidth]{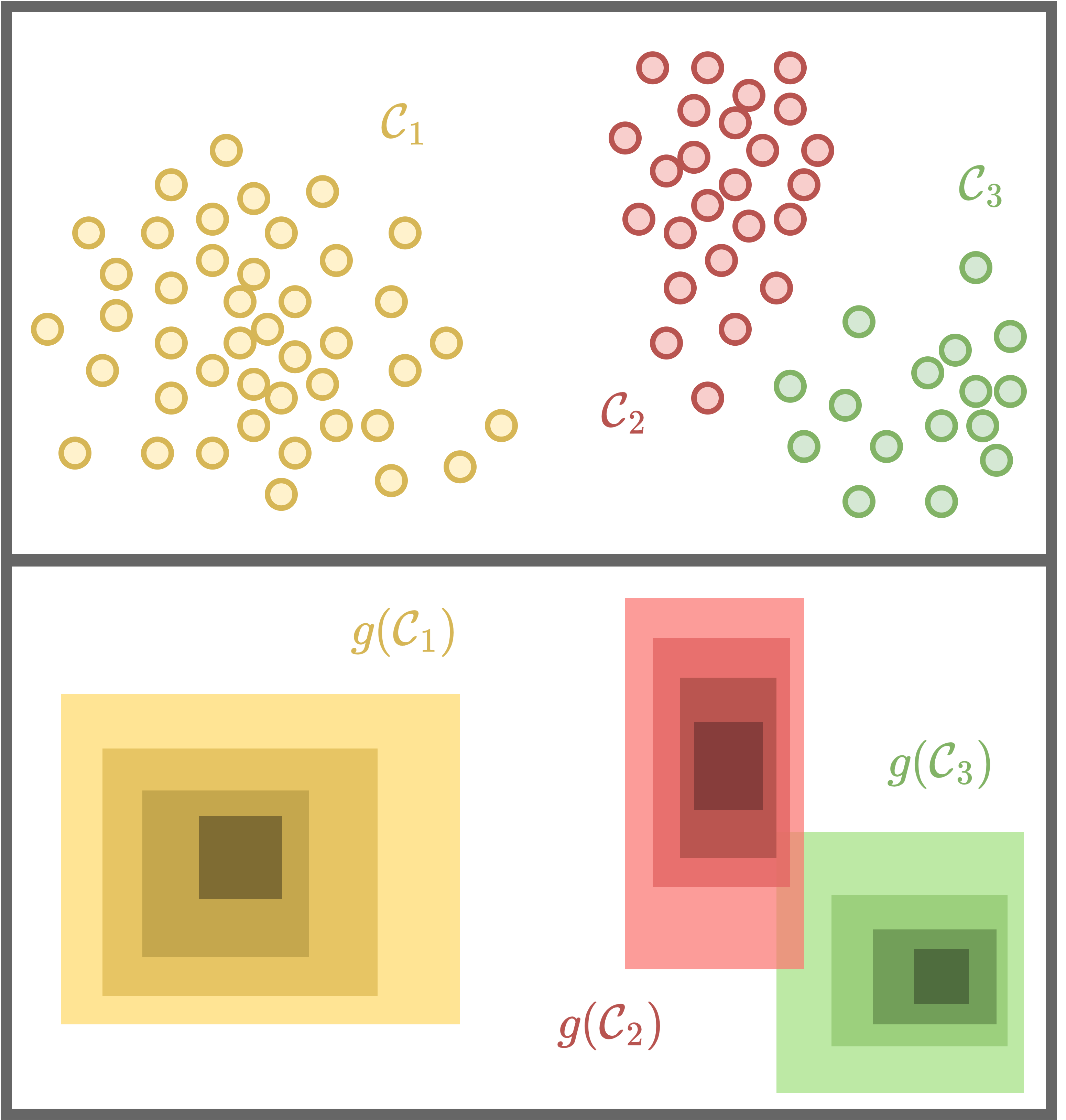}
    \caption{Summarizing clusters with quantiles. Each point represents a different PV system. In this case, the full profile length of each system is $T$=2.}
    \label{fig:quantiling}
\end{figure}

Due to the novel probabilistic nature of the entities, testing the performance of the proposed clustering method is not feasible using conventional clustering metrics such as the silhouette score. Instead, we propose a novel testing methodology that measures the representativeness of the clusters. Intuitively, we envision replacing each PV system with its respective cluster representative, which aids in (1) condensing the dataset and (2) imputing the missing values. For this, we define a cluster summarization mapping $G:\mathcal{C}\mapsto\mathbf{Y}$ resulting in a $T$-by-$Q$ matrix where its columns correspond to the quantile levels $\mathcal{Q}=\{q_i\}_{i=1}^Q$ and each row contains the quantiles obtained at timestep $t$ using the empirical distribution formed by the cluster members, i.e. $y^{(ti)}=G^{(ti)}(\mathcal{C})=\text{Quantile}_{q_i}(\{x_{u'}^{(t)} \}_{u' \in \mathcal{C}})$. 
This mapping is depicted in Fig. \ref{fig:quantiling}. 

Note that this summarization resembles a nearest-neighbors-based quantile regression. Thus, we use the quantile score
\begin{equation}\label{eq:quantile_score}
    s(\mathbf{x}_u,\mathbf{Y}_u) = \frac{1}{TQ} \sum_{t,i}^{T, Q}  \max( q_i(x^{(t)}_u - y^{(ti)}_u), (1-q_i)(y^{(ti)}_u - x^{(t)}_u))
\end{equation}
to define a dispersion score as $S^{\text{disp}}=\frac{1}{U}\sum_u s(\mathbf{x}_u,\mathbf{Y}_u)$ where $\mathbf{Y_u} = G(\mathcal{C}^u)$ and $\mathcal{C}^u$ is the label set of the cluster that the entity $u$ belongs to. The lower the dispersion score, the higher the  \textit{member-representativeness}.
We also define a leave-one-out-based sensitivity score as $S^{\text{sens}}=\frac{1}{U}\sum_u s(\mathbf{x}_u, \mathbf{Y}_{\setminus u})$ where $\mathbf{Y}_{\setminus u}=G(\mathcal{C}^u \setminus u)$ is the quantile summarization of the entity $u$'s cluster excluding itself. Essentially, this score reflects the degree of dependency of the clusters on their members individually and is desired to be low. For instance, an outlier in a cluster results in a higher sensitivity score.

\subsection{Baseline}
We employed the tilt and azimuth angles of the PV panels as a \textit{physics-based} entity representations since the temporal variations in the measurements are strongly tied to such physical parameters. These parameters are deterministic and reside in an Euclidean space. Thus, we applied k-means to these $U$ angle pairs to obtain the clusters $\{\mathcal{C}^{\text{phy}}_c\}_{c=1}^C$ for comparison.

\subsection{Hyperparameter settings}
As can be seen from Fig. \ref{fig:flow_diagram}, the proposed clustering method requires a hyperparameter setting that consists of the number of clusters ($C$), number of topics ($K$), wording granularity ($W$), and the selections of statistical distance and linkage criterion. In order to inspect the effect of these elements on the clustering performance, we experimented on a grid of hyperparameter settings. After eliminating the settings that result in at least one cluster with less than two members, we ended up with 1055 hyperparameter combinations and their respective scores. As for the quantile levels, we used $\mathcal{Q}=\{.05, .10, .25, .40, .50, .60, .75, .90, .95\}$ in this study.
\section{Results}

\begin{figure}
    \centering
    \includegraphics[width=\linewidth]{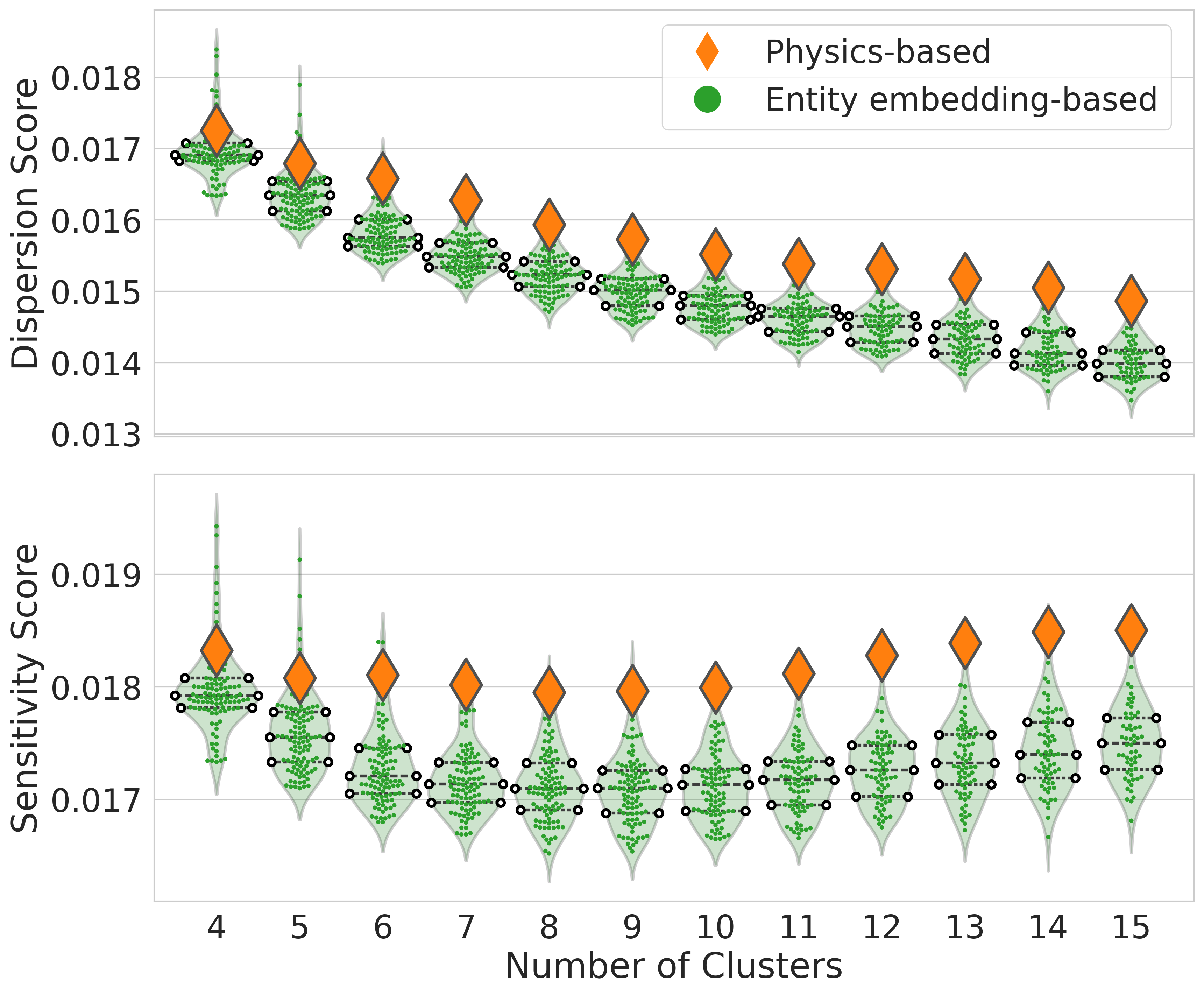}
    \caption{The effect of the number of clusters on the dispersion and sensitivity scores for physics- and entity embedding-based clustering methods. Each green dot corresponds to a distinct hyperparameter setting. The white circles on the violin plots represent the quartiles of the given setting populations.}
    \label{fig:clusters_and_methods}
\end{figure}

\begin{figure}
    \centering
    \includegraphics[width=\linewidth]{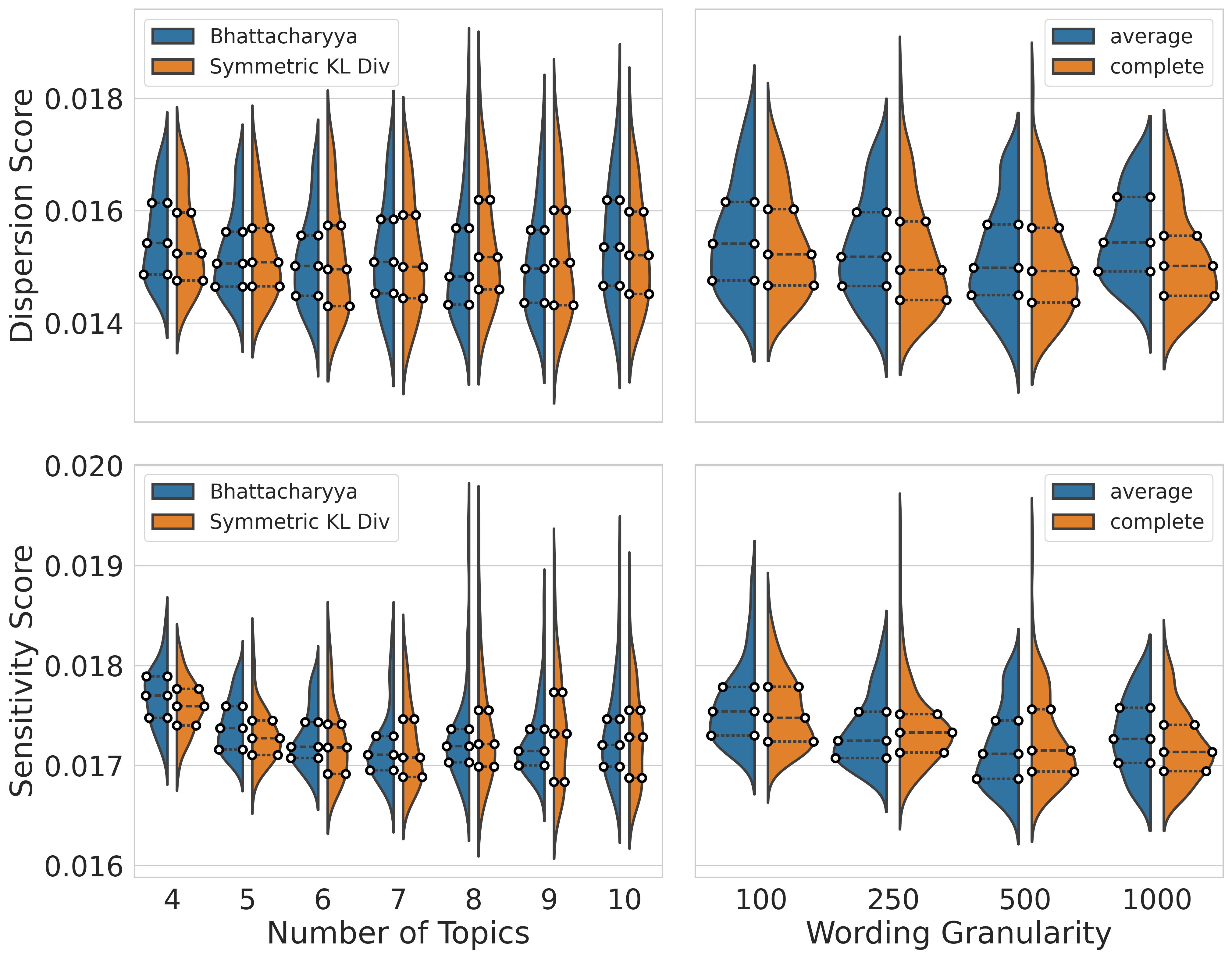}
    \caption{The effects of number of topics, wording granularity, and statistical distance and linkage criterion choices on the dispersion and sensitivity score for entity embedding-based clustering. The white circles represent the quartiles of their respective population.}
    \label{fig:ablation}
\end{figure}

\begin{figure*}
    \centering    \includegraphics[width=0.95\linewidth]{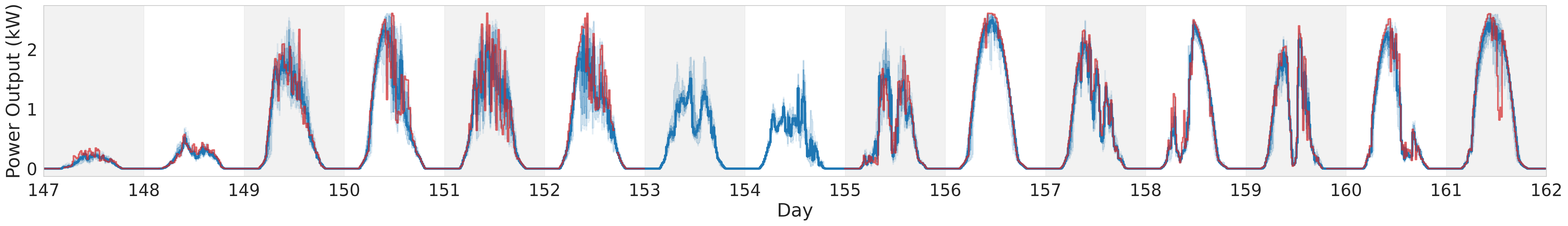}
    \caption{PV system 163's 15-day measurement profile out of 4 years (in red) and the quantile representation of the cluster it belongs to. The [153, 155] interval is completely missing for this system.}
    \label{fig:quantile_representation}
\end{figure*}

First, we investigated the effect of the number of clusters on the performance compared to the physics-based clustering. As can be seen in Fig. \ref{fig:clusters_and_methods}, our purely data-driven entity-based clustering consistently outperforms the physics-based clustering on both scores. Thus, we can deduce that there are more latent factors than PV panel angles that describe the temporal variation in the measurements, and our method implicitly captures those. Regarding the effect of the number of clusters, as expected, the dispersion score decreases with it due to the nearest-neighbor nature of the assessment. However, we see that the sensitivity score has a more convex nature since fewer (larger) clusters means inclusion of outliers, while more (smaller) clusters means higher dependency on individual members. Thus, this two-dimensional assessment provides a way to optimize the number of clusters. For this case study, we found this number as $C=8$ as the minimizer of the objective function $S^{\text{disp}}+|S^{\text{disp}}-S^{\text{sens}}|$ by using the median values of the configuration results.

We also visualized the effects of the remaining hyperparameters in Fig. \ref{fig:ablation}. We infer the following effects:
\begin{itemize}
    \item A lower \textbf{number of topics} is more robust to hyperparameter selection, yet results in a higher sensitivity score. Similar to the number of clusters, it can be chosen optimally to balance this trade-off.
    \item We observe that a higher \textbf{wording granularity} results in slightly better sensitivity score and robustness to hyperparameter settings. So, we recommend using a substantially high wording granularity.\footnote{Recall that wording granularity corresponds to the number of k-means clusters in Section \ref{subsec:wording}. So, the main bottleneck here is the computational resources (time and memory) used for the entity embedding scheme.}
    \item As for the \textbf{statistical distance} selection, we see that symmetric KL divergence is preferable for lower topic counts, while Bhattacharyya distance yields better sensitivity scores and robustness for higher counts.
    \item Lastly, the choice of \textbf{linkage criterion} appears not to substantially affect the overall performance.
\end{itemize}
Note that these hyperparameters are effective on the dispersion score only marginally. The proposed leave-one-out-based sensitivity score, on the other hand, reflects these variations better and serves as a better model selection metric.

Lastly, we visualized the cluster summarization of the cluster that PV system 163 belongs to in Fig. \ref{fig:quantile_representation}. Note that measurements for two days are completely missing. This quantile-based summarization provides an informed imputation for these days, thanks to the similarly behaving systems in the same cluster. Moreover, the overall visual alignment encourages representing the systems in the cluster with this single quantile representation for data condensation.

\section{Conclusion}

We introduce a probabilistic entity embedding‑based clustering framework that represents each PV system as a probability distribution, thereby encoding both its characteristic behavior and the uncertainty due to data gaps, and then applies statistical distance measures alongside agglomerative clustering to form coherent groups of systems. We validated our approach on a multi‑year dataset of 175 residential rooftop PV power measurements, showing that it delivers notably more representative and robust cluster profiles than a physics‑informed baseline built on tilt and azimuth angles. By assessing clusters via a quantile‑based dispersion metric and a leave‑one‑out sensitivity metric, our method not only facilitates dataset condensation and missing-value imputation but also yields stable clusters less dependent on individual outliers. Additionally, a systematic exploration of hyperparameter choices provides practical guidance for tuning the model to balance representativeness and robustness.

As for the future work, we aim to apply our method to different multi-site power systems data, such as smart meters and feeders. The absence of PV panel angles-like baseline representations makes it harder to assess the performance improvement for these domains. Thus, we desire to expand this study with downstream applications like optimal market bidding and state estimation to evaluate the representativeness of clusters by comparing against the original profiles.

\bibliographystyle{IEEEtran}
\bibliography{references.bib}

\begin{thebibliography}{1}
\providecommand{\url}[1]{#1}
\csname url@samestyle\endcsname
\providecommand{\newblock}{\relax}
\providecommand{\bibinfo}[2]{#2}
\providecommand{\BIBentrySTDinterwordspacing}{\spaceskip=0pt\relax}
\providecommand{\BIBentryALTinterwordstretchfactor}{4}
\providecommand{\BIBentryALTinterwordspacing}{\spaceskip=\fontdimen2\font plus
\BIBentryALTinterwordstretchfactor\fontdimen3\font minus \fontdimen4\font\relax}
\providecommand{\BIBforeignlanguage}[2]{{%
\expandafter\ifx\csname l@#1\endcsname\relax
\typeout{** WARNING: IEEEtran.bst: No hyphenation pattern has been}%
\typeout{** loaded for the language `#1'. Using the pattern for}%
\typeout{** the default language instead.}%
\else
\language=\csname l@#1\endcsname
\fi
#2}}
\providecommand{\BIBdecl}{\relax}
\BIBdecl

\bibitem{visser2024probabilistic}
L.~Visser, T.~AlSkaif, and W.~van Sark, ``Probabilistic solar power forecasting: An economic and technical evaluation of an optimal market bidding strategy,'' \emph{Applied Energy}, vol. 370, p. 123573, 2024.

\bibitem{strezoski2022integration}
L.~Strezoski, H.~Padullaparti, F.~Ding, and M.~Baggu, ``Integration of utility distributed energy resource management system and aggregators for evolving distribution system operators,'' \emph{Journal of Modern Power Systems and Clean Energy}, vol.~10, no.~2, pp. 277--285, 2022.

\bibitem{panigrahi2020grid}
R.~Panigrahi, S.~K. Mishra, S.~C. Srivastava, A.~K. Srivastava, and N.~N. Schulz, ``Grid integration of small-scale photovoltaic systems in secondary distribution network—a review,'' \emph{IEEE Transactions on Industry Applications}, vol.~56, no.~3, pp. 3178--3195, 2020.

\bibitem{visser2022open}
L.~R. Visser, B.~Elsinga, T.~A. AlSkaif, and W.~G. Van~Sark, ``Open-source quality control routine and multi-year power generation data of 175 pv systems,'' \emph{Journal of Renewable and Sustainable Energy}, vol.~14, no.~4, 2022.

\bibitem{bolat2024guide}
K.~B{\"o}lat and S.~Tindemans, ``Guide-{VAE}: Advancing data generation with user information and pattern dictionaries,'' \emph{arXiv preprint arXiv:2411.03936}, 2024.

\bibitem{kaufman2009finding}
L.~Kaufman and P.~J. Rousseeuw, \emph{Finding groups in data: an introduction to cluster analysis}.\hskip 1em plus 0.5em minus 0.4em\relax John Wiley \& Sons, 2009.

\bibitem{blei2003latent}
D.~M. Blei, A.~Y. Ng, and M.~I. Jordan, ``Latent dirichlet allocation,'' \emph{Journal of machine Learning research}, vol.~3, no. Jan, pp. 993--1022, 2003.

\bibitem{joram_soch_2025_14646799}
\BIBentryALTinterwordspacing
J.~Soch \emph{et~al.}, ``Statproofbook/statproofbook.github.io: Statproofbook 2024,'' Jan. 2025. [Online]. Available: \url{https://doi.org/10.5281/zenodo.14646799}
\BIBentrySTDinterwordspacing

\bibitem{rauber2008probabilistic}
T.~W. Rauber, T.~Braun, and K.~Berns, ``Probabilistic distance measures of the dirichlet and beta distributions,'' \emph{Pattern Recognition}, vol.~41, no.~2, pp. 637--645, 2008.

\end{thebibliography}

\end{document}